\documentclass[twoside]{article}

\usepackage[accepted]{aistats2020}
\usepackage{microtype}
\usepackage{graphicx}
\usepackage{tabularx,ragged2e,booktabs,caption}
\usepackage{esvect}
\usepackage{booktabs}
\usepackage{amsmath, amssymb}
\usepackage{bm}
\usepackage{amsthm}
\DeclareMathOperator*{\argmax}{arg\,max}

\usepackage{multirow}
\usepackage{diagbox}
\usepackage{mathtools}
\usepackage{graphicx}
\usepackage{subfig}
\usepackage{wrapfig}
\usepackage{esvect}
\usepackage{xcolor}
\usepackage{algorithm}
\usepackage{algorithmic}
\usepackage{wrapfig}

\usepackage{textcomp}

\usepackage{ragged2e}

\usepackage[utf8]{inputenc} 
\usepackage[T1]{fontenc}    
\usepackage{hyperref}       
\usepackage{url}            
\usepackage{booktabs}       
\usepackage{amsfonts}       
\usepackage{nicefrac}       
\usepackage{microtype}      
\usepackage{dblfloatfix}    
\usepackage{subfloat}       
\usepackage{pifont}
\newcommand{\cmark}{\ding{51}}
\newcommand{\xmark}{\textcolor{lightgray}{\ding{55}}}

\usepackage{caption}
\usepackage{enumitem}
\usepackage{balance}

\usepackage{multicol}
\usepackage{longtable}

\newcommand{\bftab}{\fontseries{b}\selectfont}

\begin{document}

%
\runningtitle{Stepwise Model Selection for Sequence Prediction\\via Deep Kernel Learning}

%

\twocolumn[

\aistatstitle{Stepwise Model Selection for Sequence Prediction\\via Deep Kernel Learning}
\vspace{-0.25cm}
\aistatsauthor{Yao Zhang \And Daniel Jarrett \And Mihaela van der Schaar  }

\aistatsaddress{University of Cambridge \And University of Cambridge \And University of Cambridge, UCLA\\ The Alan Turing Institute  } ]
\vspace{-1cm}
\allowdisplaybreaks

\begin{abstract}\vspace{-0.5em}

An essential problem in automated machine learning (AutoML) is that of model selection. A unique challenge in the sequential setting is the fact that the optimal model \textit{itself} may vary over time, depending on the distribution of features and labels available up to each point in time.
In this paper, we propose a novel Bayesian optimization (BO) algorithm to tackle the challenge of model selection in this setting. This is accomplished by treating the performance
at each time step as its own black-box function. In order to solve the resulting multiple black-box function optimization problem \textit{jointly} and \textit{efficiently}, we exploit potential correlations among black-box functions using deep kernel learning (DKL). To the best of our knowledge, we are the first to formulate the problem of \textit{stepwise} model selection (SMS) for sequence prediction, and to design and demonstrate an efficient joint-learning algorithm for this purpose. Using multiple real-world datasets, we verify that our proposed method outperforms both standard BO and multi-objective BO algorithms on a variety of sequence prediction tasks.

\end{abstract}\vspace{-0.5em}\section{Introduction}

Model selection is a central concern in automated machine learning (AutoML). Techniques using Bayesian optimization (BO) have proven popular and effective for this purpose \cite{snoek2012practical}, and have been extended to incorporate trade-offs between multiple objectives \cite{hernandez2016predictive, picheny2015multiobjective, shah2016pareto}, as well as accommodating transfer across multiple tasks \cite{perrone2018scalable, SwerskySA13, Yehong17}. In this paper, we focus on Bayesian optimization for model selection in the \mbox{\textit{sequence prediction}} setting\textemdash that is, where the underlying task is to emit predictions $e_{t}$ at every step $t$ given a sequence of observations $\{\mathbf{o}_t\}_{t=1}^{T}$ as input. Instead of studying the dynamics between or across different \textit{tasks}, we concentrate on how the optimal model itself (for a fixed task) may vary \textit{over time}, depending on the distribution of features and labels in the data available at each time.

Note that there are two different senses of changes ``over time''. The first concerns distribution shifts \textit{across} successive batches of (static) data: If the data-generating process evolves across multiple datasets, existing models trained on prior data may require updating as new batches become available\textemdash that is, in order to continue to generalize well.
This \textit{sequential} process can be assisted for instance by hyperparameter transfer learning \cite{perrone2018scalable}, and is not the focus of this work.
The second type is more subtle, and is the motivation for our work: There may be temporal distribution shifts \textit{within} the same dataset that we are attempting to learn from. Unlike in sequential hyperparameter transfer learning, here all the data we need is already available, and learning can in principle be done \textit{jointly} across all time steps. Temporal distribution shifts often arise in healthcare, and can happen on both the individual and population level. As an example of the former, the risk factors for an adverse outcome at the beginning of a patient's hospital stay may generally be different from those that govern their condition towards the end of the episode. As an example of the latter, as a medical study progresses over time, the distribution of registered patients and their treatments and outcomes may undergo a shift. As noted in \cite{oh2019relaxed, wiens2016patient}, while such phenomena are common in the medical setting, they are rarely addressed explicitly by current \textit{single-model} techniques\textemdash potentially giving rise to suboptimal prediction performance.

In this paper, we develop an automated technique for \textit{stepwise} model selection (SMS) over time, thereby tackling the challenge of
optimal models evolving throughout a dataset.
We propose a novel BO algorithm for
SMS,
treating the prediction performance at each time step as its own black-box function. To solve the resulting multiple black-box function optimization problem \textit{jointly} and \textit{efficiently}, we exploit correlations among black-boxes via deep kernel learning (DKL). Using real-world datasets in healthcare, we verify that our method outperforms both standard BO and multi-objective BO algorithms on a variety of sequence prediction tasks. \mbox{To the best} of our knowledge, we are the first to formulate the problem of stepwise model selection
for sequence prediction, and to design and demonstrate an efficient algorithm for this purpose. Our technique \mbox{contributes} to AutoML in developing powerful seq- uence models while keeping the human out of the loop.

\section{Problem Formulation} 

To establish notation, we first introduce the underlying sequence prediction task, and formalize the stepwise model selection problem that we address in this paper.

\textbf{Sequence Prediction}. Let $\mathbf{o}_t\in{\mathbb{R}^{d}}$ denote (observed) input variables, and $e_t\in{\mathbb{R}}$ the (emitted) output variable, where $t\in\{1,...,T\}$ in sequences of up to length $T$. At every time $t$, the underlying task is to predict the label $e_t$ on the basis of the sequence of observations available up until time $t$: $(\mathbf{o}_{\tau})_{\tau=1}^{t}$. To this end, we are given a finite dataset $\mathcal{D}=\{(\mathbf{o}_{i,t},e_{i,t})_{t=1}^{T}\}_{i=1}^{I}$
for training purposes, where individual sequences are indexed by $i\in\{1,...,I\}$ in a dataset with $I$ sequences. Let $\mathbb{X}$ denote the space of \textit{hyperparameters} for such sequence prediction models,
including discrete and continuous variables that configure architectures and training. For example, the hyperparameter space of a recurrent neural network (RNN) may include\textemdash among others\textemdash the size of the hidden state,
dropout rate, and
coefficient on weight-decay. (If we were to consider different classes of models entirely, e.g.
GRUs vs. LSTMs,
this can also be accommodated via additional categorical dimensions).

\textbf{Stepwise Model Selection}. In standard Bayesian optimization, the task is to minimize (or maximize) some black-box function $f : \mathbb{X} \to \mathbb{R}$. Let $a_f : \mathbb{X} \to \mathbb{R}$ denote the acquisition function, which captures the utility of evaluating $f$ at $\mathbf{x}\in \mathbb{X}$. At each BO iteration, we use $a_{f}$ to determine the next point to evaluate, and the goal is to find the global minimizer (or maximizer) of $f$ after the fewest iterations. Concretely, let $\mathcal{D}_{\leq t}=\{(\mathbf{o}_{i,\tau},e_{i,\tau})_{\tau=1}^{t}\}_{i=1}^{I}$ give the filtration of the full dataset $\mathcal{D}$ with respect to time $t$, and let $\mathcal{L}_{t}$ denote the validation performance metric of interest (e.g. likelihood of the data, area under the receiver operating characteristic, etc.) for time step $t$. The (conventional) \textit{single-model} approach for sequence prediction is to find a single maximizer $\mathbf{x}^{*}$ that is used for \textit{all} time steps $t$,
\begin{equation}
\label{eq:obj2}
\mathbf{x}^{*}\in\argmax_{\mathbf{x}\in\mathbb{X}}
\sum_{t=1}^{T}\mathcal{L}_{t}(\mathbf{x},\mathcal{D}^{\text{train}},\mathcal{D}_{\leq t}^{\text{valid}})
\end{equation}
where superscripts on $\mathcal{D}$ denote training and validation splits. Defining $f(\mathbf{x})=\sum_{t=1}^{T}\mathcal{L}_{t}(\mathbf{x},\mathcal{D}^{\text{train}},\mathcal{D}_{\leq t}^{\text{valid}})$ gives us the black-box function to be optimized using BO.

In this paper, we extend this formulation to accommodate the SMS problem\textemdash that is, of selecting the best sequence prediction model for \textit{each} time step. To this end, we treat the prediction performance
at each step $t\in\{1,...,T\}$ as its own a black-box function $f_{t}$. Our objective is to find the best $\mathbf{x}_{t}^{*}$ that maximizes each $f_{t}$; in other words, we want the set of \textit{stepwise} maximizers,
\begin{equation}
\label{eq:obj}
\{\mathbf{x}_{t}^{*}\}_{t=1}^{T}\in\argmax_{\{\mathbf{x}_{t}\}_{t=1}^{T}\in\mathbb{X}^{T}}
\sum_{t=1}^{T}\mathcal{L}_{t}(\mathbf{x}_{t},\mathcal{D}^{\text{train}},\mathcal{D}_{\leq t}^{\text{valid}})
\end{equation}
where for brevity we use $\mathbb{X}^{T}$ to denote $\prod_{t=1}^T\mathbb{X}$. Defining $f_{t}(\mathbf{x}_{t})=\mathcal{L}_{t}(\mathbf{x}_{t},\mathcal{D}^{\text{train}},\mathcal{D}_{\leq t}^{\text{valid}})$ for $t\in\{1,...,T\}$ then gives $T$ black-box functions to be optimized using BO.

\textbf{Multiple Black-Boxes}. Two points require emphasis. The first concerns the problem (SMS), and the second motivates our solution (DKL). First, the $T$ black-box functions are in general \textit{distinct}. In the presence of potential distribution shifts, it is highly unlikely that the optimizer for all $t$ will be the same exact model. For instance, the optimal RNN for the first 24 hours of an ICU physiological stream may require little recurrent memory as the typical patient is very stable; yet as more patients enter deteriorating states over time, we may require more complex hidden states that can better capture both short and long-range patterns. Reducing the SMS problem in (\ref{eq:obj2}) to a single black-box problem as in (\ref{eq:obj}) may be overly constraining; we will observe examples of this in our experiments later in Section \ref{sect:experiments}.

Second, however, the $T$ black-box functions are in general \textit{not independent}. For one, we expect a given model's performance to be correlated across time\textemdash especially between neighboring steps. Furthermore, instead of obtaining a single point $(\mathbf{x}_{n},y_{n}=f(\mathbf{x}_{n}))$ per acquisition step $n$, here we obtain a total of $T$ points $\{(\mathbf{x}_{n,t},y_{n,t})\}_{t=1}^{T}$ per acquisition; these are obtained \textit{simultaneously}, since we can observe a model's performance for all times $t$ for every evaluation (i.e. with a single pass through $\mathcal{D}$). Denote by $\mathcal{A}_t =(\mathbf{X}_t, \mathbf{y}_t)$ the acquisition set for time $t$, where the $n$-th row of $\mathbf{X}_t$ corresponds to $\mathbf{x}_{n,t}$, and the $n$-th entry of $\mathbf{y}_{t}$ to $y_{n,t}$. The number of rows in $\mathbf{X}_t$ equals $N$, the total number of acquisitions\textemdash the same for all functions $f_t$. In our proposed solution, we will leverage the correlations between time steps $t$ to jointly optimize all $f_1,...,f_T$, as well as acquiring models via a soft policy prioritizing black-boxes with the highest expected improvement.


\begin{figure*}[!t]
\begin{center}
\centerline{\hspace*{-0.325cm}\includegraphics[width=0.90\linewidth]{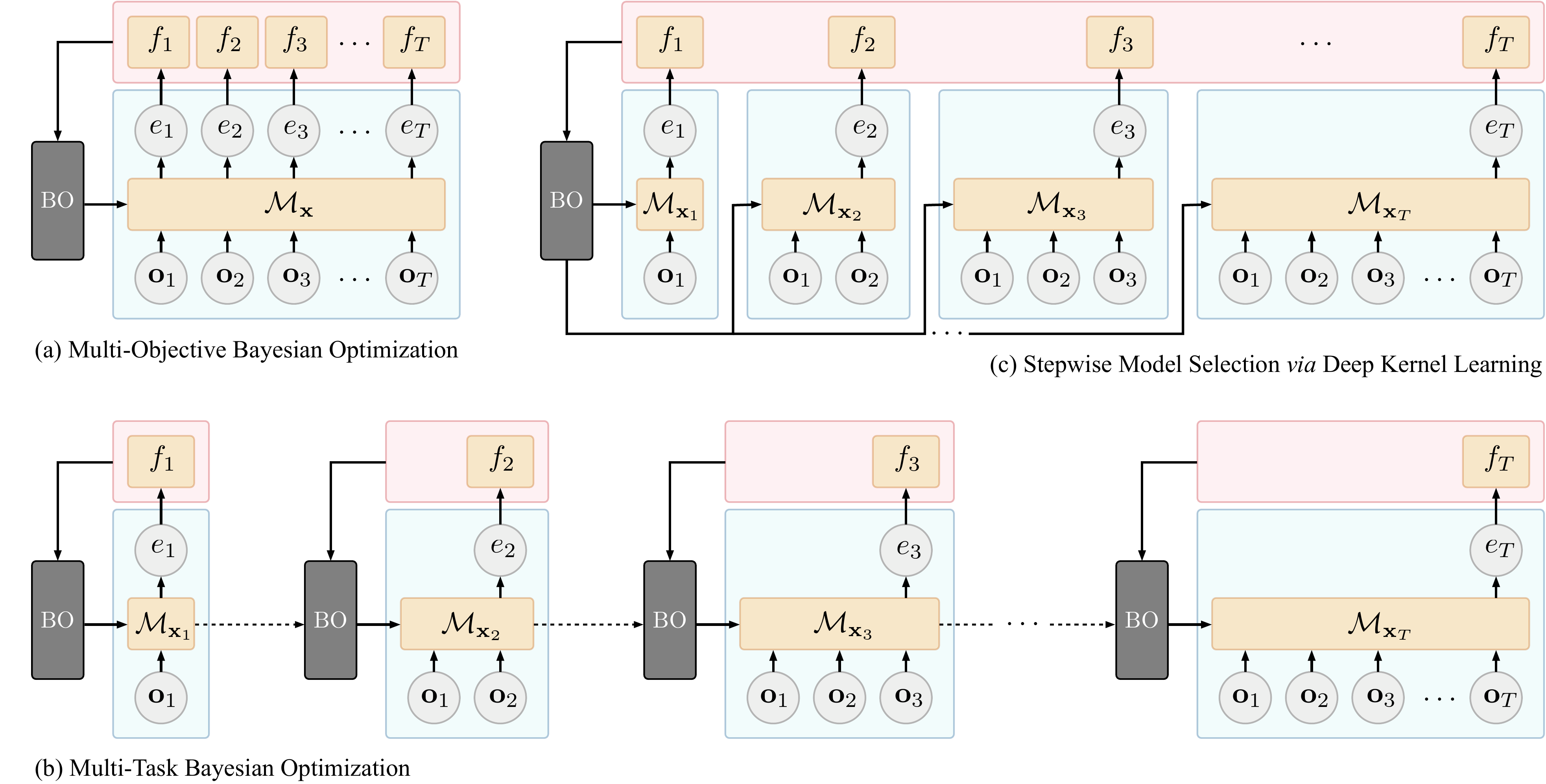}}
\caption{Comparison of related methods in the context of model selection for sequence prediction. Each $\mathcal{M}_{\mathbf{x}}$ indicates a model (hyper-)parameterized by $\mathbf{x}$. (a) Multi-objective Bayesian optimization, which is constrained to learn a single model for all time steps. (b) Multi-task Bayesian optimization, which can be applied \textit{sequentially} across time steps. (c) Our proposed technique for stepwise model selection via deep kernel learning, which \textit{jointly} learns all models for all time steps.}
\label{fig:time_model}
\end{center}
\vskip -0.3in
\end{figure*}

\begin{figure}[!b]
\begin{center}
\vskip -0.20in
\centerline{\includegraphics[width=0.65\columnwidth]{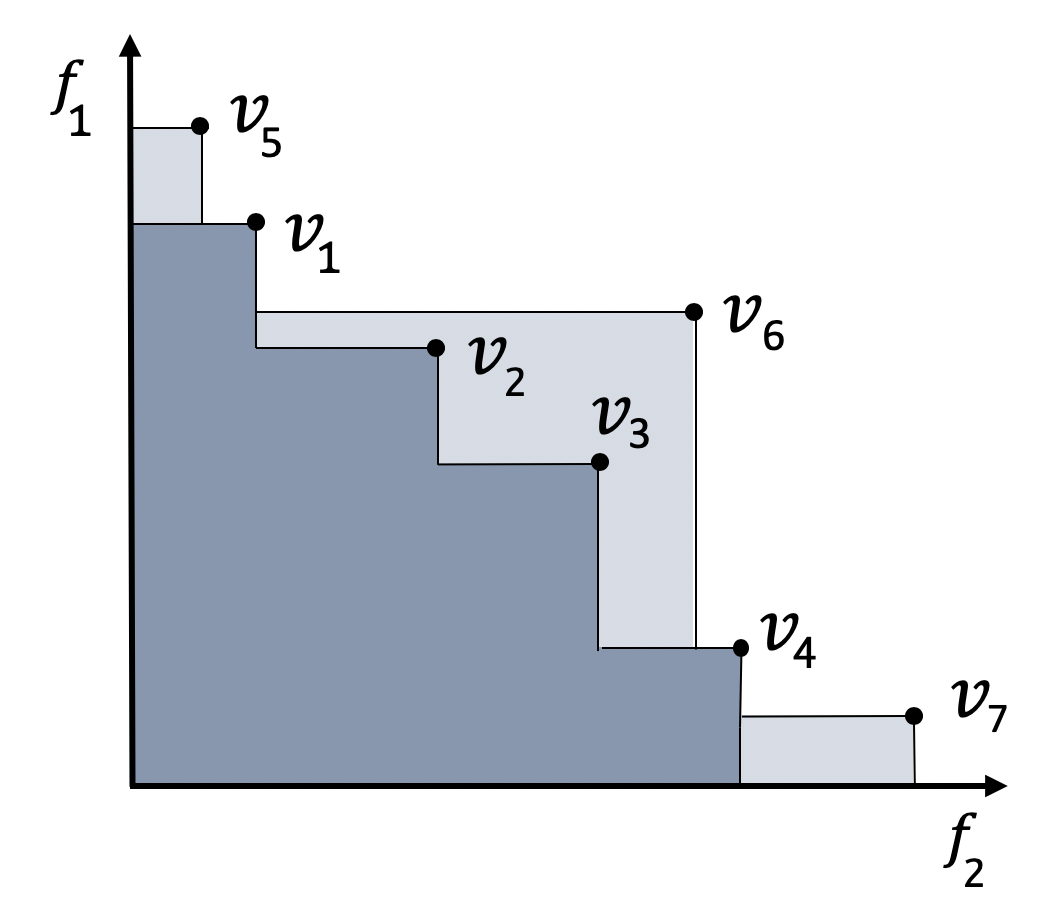}}
\vskip -0.12in
\caption{Example of Pareto frontier estimates with two objectives. In the MOBO setting we would often find $v_6$ the most attractive \textit{single} solution, while in the SMS setting we would be \textit{simultaneously} interested in both $v_5$ and $v_7$.}
\label{fig:Pareto_front}
\end{center}
\vskip -0.21in
\end{figure}

\section{Related Work}

We take on the problem of selecting sequence predict- ion models for each time step, casting this as a multiple black-box optimization problem. As such, our work bears some resemblance to multi-objective Bayesian optimization, and to multi-task Bayesian optimization.

\textit{Multi-Objective} Bayesian optimization (MOBO) \cite{emmerich2008computation,hernandez2016predictive,knowles2006parego,picheny2015multiobjective,ponweiser2008multiobjective, shah2016pareto,zitzler1999multiobjective} deals with optimizing multiple objectives in a trade-off relationship. Consider two objectives $f_{1},f_{2}$ as in Figure \ref{fig:Pareto_front}. Suppose that, after a certain number of BO iterations, our current best estimate of the Pareto frontier is given by points $v_1$ through $v_4$ (the feasible region is shaded in dark). Further suppose that, as more points are sampled, the frontier is pushed outward by additional points $v_5$ through $v_7$. In the MOBO setting, $v_6$ would often provide the most attractive trade-off between the two objectives\textemdash for instance, based on hyper-volume gain. In contrast, our goal in the SMS setting is to find an optimal model $\mathbf{x}_{t}^{*}$ at \textit{each} time $t$; importantly, there is no trade-off relationship\textemdash each such $\mathbf{x}_{t}^{*}$ does \textit{not} need to be optimal for any other time step. In this example, our primary interests are therefore in $v_5$ (for $f_1$) and $v_7$ (for $f_2$), but not $v_6$. If (for whatever reason) we were constrained to select a single prediction model for all time steps, then the SMS problem would be reduced to MOBO.

\newcolumntype{A}{>{\centering\arraybackslash}m{1.5cm}}
\newcolumntype{B}{>{\centering\arraybackslash}m{4.4cm}}
\newcolumntype{C}{>{\centering\arraybackslash}m{1.4cm}}
\newcolumntype{D}{>{\centering\arraybackslash}m{2.2cm}}
\newcolumntype{E}{>{\centering\arraybackslash}m{2.9cm}}
\newcolumntype{F}{>{\centering\arraybackslash}m{2.8cm}}

\setlength\tabcolsep{4.6pt}
\begin{table*}[t]
\caption{Comparison of related methods in the context of model selection for sequence prediction. $^{1}$Note that correlations \textit{within} black-box functions are exploited in all BO methods; SMS-DKL additionally exploits correlations \textit{across} functions. $^{2}$Some MOBO methods achieve this, but they are \textit{not} scalable to problems with large numbers of objectives (see Section \ref{sect:experiments}).}
\vspace{-1em}
\begin{small}
\begin{center}
\begin{tabular}{ABCDEF}
\toprule
& {\scriptsize Optimization Problem} & {\scriptsize Number of Optimizers} & {\scriptsize Optimize All Functions Jointly} & {\scriptsize Exploit Correlations among All Functions$^{1,2}$} & {\scriptsize Prioritize which Functions to Optimize} \\
\midrule
MOBO                       & \raisebox{1.2em}{$\mathbf{x}^{*}\in\displaystyle\argmax_{\mathbf{x}\in\mathbb{X}}\sum_{t=1}^{T}f_{t}(\mathbf{x})$} & $1$ & \cmark & \xmark & \xmark \\
\raisebox{-0.5em}{MTBO}    & \mbox{for $t\in\{1,...,T\}$:} \mbox{$\mathbf{x}_{t}^{*}\in\displaystyle\argmax_{\mathbf{x}_{t}\in\mathbb{X}}f_{t}(\mathbf{x}_{t})$} & \raisebox{-0.5em}{$T$} & \raisebox{-0.5em}{\xmark} & \raisebox{-0.5em}{\cmark} & \raisebox{-0.5em}{\xmark} \\
SMS-DKL                    & \raisebox{1.2em}{$\{\mathbf{x}_{t}^{*}\}_{t=1}^{T}\in\displaystyle\argmax_{\{\mathbf{x}_{t}\}_{t=1}^{T}\in\mathbb{X}^{T}}\sum_{t=1}^{T}f_{t}(\mathbf{x}_{t})$} & $T$ & \cmark & \cmark & \cmark \\
\bottomrule
\end{tabular}
\vspace{-1.5em}
\end{center}
\label{tab:related}
\end{small}
\end{table*}

\textit{Multi-Task} Bayesian Optimization (MTBO) \cite{perrone2018scalable,SwerskySA13,Yehong17,zhang2019lifelong} deals with transferring knowledge gained from previous optimizations to new tasks, such that subsequent optimizations are more efficient. This setting applies, for example, to the problem where successive batches of (static) data are available or accumulated over time, such that prior trained models may require retraining. Of course, the SMS problem can be (naively) reduced to an MTBO problem\textemdash that is, we can optimize all models $f_t$ \textit{sequentially}, for instance by using optimizations of black-box functions $f_1,...,f_{t-1}$ to warm-start the optimization for $f_t$. However, this approach is of little practical interest. Evaluating deep learning models on large datasets is expensive, and in practice we have a limited computational budget\textemdash we are interested in finding a good model after a set number of BO evaluations (depending on the dimension of the hyperparameter space). Conducting SMS by reduction to MTBO requires $T$ separate BO procedures in a sequence, and it is unclear how to allocate evaluations among these subproblems while keeping the human out of the loop. In addition, in contrast to the \textit{joint} approach of our proposed solution (which involves a single BO procedure for all $f_t$), MTBO does not take full advantage of information from all acquisition functions.

In this paper, we take on the SMS problem for sequence prediction models by optimizing all functions $f_1,...,f_{T}$ \mbox{\textit{at the same time}}. In contrast to MOBO, we are not constrained by trade-offs between competing objectives. And in contrast to MTBO, our goal is to take full advantage of the potential correlations among black-box functons $f_t$, as well as information from the acquisition functions, by learning the models for all steps jointly. See Figure \ref{fig:time_model} and Table \ref{tab:related} for a comparison of MBTO, MOBO, as well as our proposed approach\textemdash SMS-DKL\footnote{An implementation of SMS-DKL is available at https://bitbucket.org/mvdschaar/mlforhealthlabpub.}.

\section{SMS via Deep Kernel Learning}

We now develop our proposed technique for the SMS problem: a novel BO algorithm that uses deep kernel learning (DKL) to solve the multiple black-box optimization problem jointly and efficiently. Recall that each black-box function $f_t$ corresponds to the validation performance $\mathcal{L}_t$. This depends on the filtration $\mathcal{D}_{\leq t}$ and the selected model $\mathbf{x}_{t}$; accordingly, we expect the similarity among black-box functions $f_1,...,f_T$ to be explained by interactions between filtrations and models.
Kernels are often deployed as measures of similarity: They are used in support vector machines and gaussian processes to measure the similarity $K(\mathbf{v},\mathbf{v}^{\prime})$ between two vectors (using an inner product in a transformed space); they can also be used to measure the similarity $K(p,p^{\prime})$ between two distributions (using the sample average of an inner kernel, for instance) \cite{muandet2012learning}.
Here, we propose a deep kernel learning method designed to measure the similarity between two \textit{filtration-and-model} tuples. Let $(\mathcal{D}_{\leq t},\mathbf{x}_t)$ and $(\mathcal{D}_{\leq t^{\prime}},\mathbf{x^{\prime}}_{t^{\prime}})$ be two such pairs; we will allow their similarity to be captured by a learned kernel parameterized by a neural network.

Section \ref{sect:deep_kernels} describes how to capture similarities between black-boxes, Section \ref{sect:learning} covers learning and inference, and Section \ref{sect:acquisition} provides the acquisition method.

\subsection{Deep Kernels}\label{sect:deep_kernels}

\begin{figure}[t]
\centering
\includegraphics[width=1.0\linewidth]{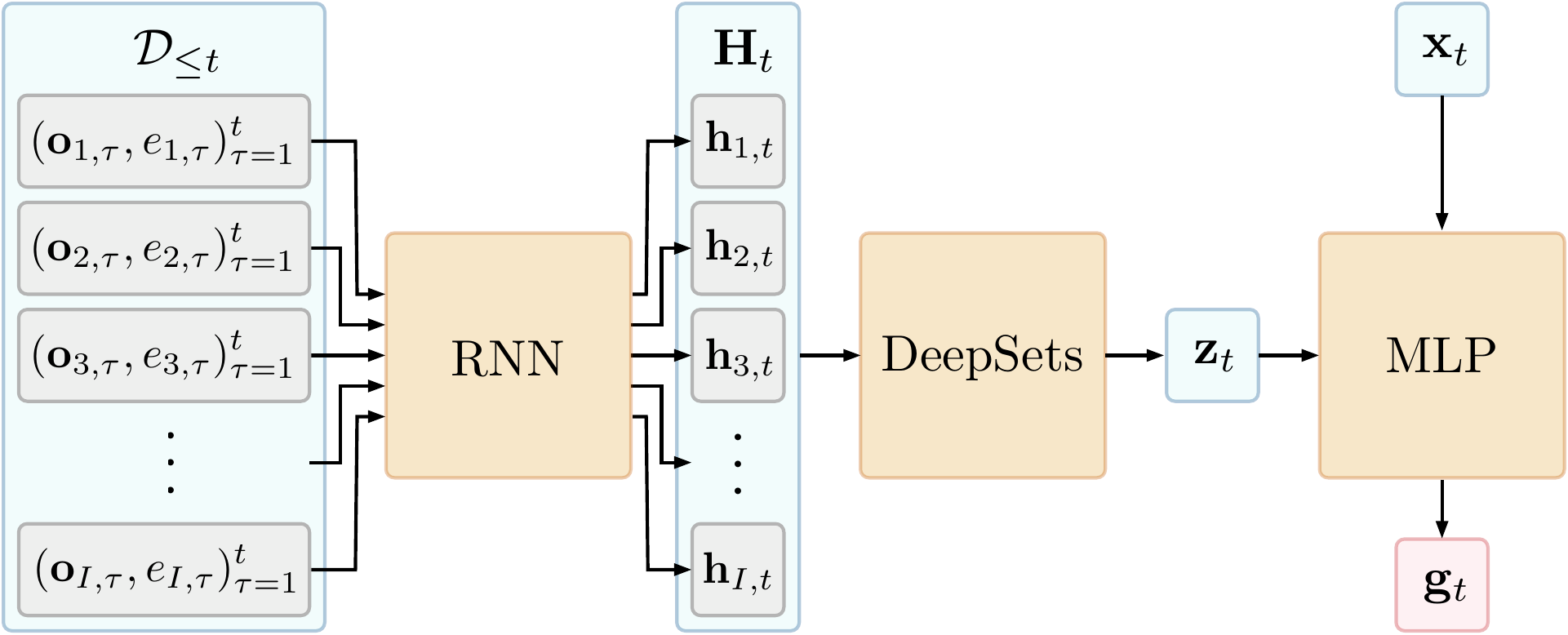}
\vspace{-1.5em}
\caption{Architecture of DKL network.}
\vspace{-1.0em}
\label{fig:net}
\end{figure}

\textbf{Vector Embedding}. We start by transforming tuples $(\mathcal{D}_{\leq t},\mathbf{x}_t)$ into \textit{fixed-length} vector representations $\mathbf{g}_t$; these are the feature maps that will subsequently be used by the kernel to measure similarities between such tuples. This is accomplished through a neural network that consists of three components. First, an RNN learns a per-instance representation of $\mathcal{D}_{\leq t}$\textemdash that is, of $(\mathbf{o}_{i,\tau},e_{i,\tau})_{\tau=1}^{t}$ for all $i\in\{1,...,I\}$. Denote by $\textbf{h}_{i,t}$ the embedding for instance $i$; the result of this step is therefore given by the matrix $\textbf{H}_{t}$. Second, we pass $\textbf{H}_{t}$ through a DeepSets network \cite{zaheer2017deep} in order to obtain a \textit{permutation-invariant} embedding, which we denote by $\mathbf{z}_t$. This is important: we want an embedding that is independent of how the individual rows are ordered in $\textbf{H}_{t}$. Third, the vectors $\mathbf{z}_t,\mathbf{x}_t$ are concatenated and fed into a multilayer perceptron (MLP) that generates the final feature map $\mathbf{g}_t$.
See Figure \ref{fig:net} for a block-diagram.

In DKL, $\mathbf{z}_t$ can be interpreted by analogy to a set of informative statistics on $\mathcal{D}_{\leq t}$; 
then the MLP component simply operates as a standard deep kernel machine \cite{al2016learning,wilson2016deep} that takes vectors $\mathbf{z}_t \oplus \mathbf{x}_t$ as input. Importantly, instead of relying on predefined measures of the data or handcrafted statistical meta-features to explain model performance \cite{bardenet2013collaborative, feurer2015efficient, pfahringer2000meta}, here we allow informative measures to be flexibly learned by way of neural networks.

\setlength{\abovedisplayskip}{6pt}
\setlength{\belowdisplayskip}{6pt}

\textbf{Kernel Construction}. We now show how to use the feature map $\mathbf{g}_t$ to construct a deep kernel. Recall that the acquisition set for time $t$ is given by $\mathcal{A}_t =(\mathbf{X}_t, \mathbf{y}_t)$; here $\mathbf{X}_t$ contains $N$ rows (and $\mathbf{y}_t$ contains $N$ entries), where $N$ is the number of acquisitions made so far. Let $\mathbf{G}_t$ denote the $N\times D$ matrix where the $n$-th row is given by ($D$-dimensional) $\mathbf{g}_{n,t}$, the feature map corresponding to $\mathbf{x}_{n,t}$. The deep linear kernel machine is constructed by performing a Bayesian linear regression on $\mathbf{g}_t$\textemdash that is, by marginalizing out the output layer $\mathbf{w}_t$ on the top of the feature map $\mathbf{g}_t$, with respect to the posterior distribution of $\mathbf{w}_t$. The likelihood function is as follows,
\begin{equation}\label{equ:likelihood}
\begin{split}
P(\mathbf{y}_t|\mathbf{X}_t,\mathbf{w}_t, \mathbf{\Theta}_t ) =\prod_{n=1}^{N}\mathcal{N}(y_{n,t};  \mathbf{G}_{t} \mathbf{w}_{t},\beta_{t}^{-1})
\end{split}
\end{equation}
where $\beta_t$ is the precision parameter. The prior distribution is $P(\mathbf{w}_{t}|\lambda_t)=\mathcal{N}(\mathbf{0},\lambda_{t}^{-1}\mathbf{I}_{D\times D})$ with precision parameter $\lambda_t$, which leads to the posterior distribution,
\begin{equation} \label{eq:h_post}
P(\mathbf{w}_t|\mathcal{A}_{t},\mathbf{\Theta}_t)= \mathcal{N}(\mathbf{m}_{\mathbf{w}_t},\mathbf{K}_{\mathbf{w}_t}^{-1})
\end{equation}
where the set $\mathbf{\Theta}_t$ collects all neural network parameters in the DKL architecture as well as $\lambda_t$ and $\beta_t$, and the mean function $\mathbf{m}_{\mathbf{w}_t}$ and kernel $\mathbf{K}_{\mathbf{w}_t}$ are as follows,
\begin{equation*}
\mathbf{m}_{\mathbf{w}_t}=\frac{\beta_t}{\lambda_t} \mathbf{K}_{\mathbf{w}_t}^{-1}\mathbf{G}_t^{\top}\mathbf{y}_t,
\ \ \ \mathbf{K}_{\mathbf{w}_t} =\frac{\beta_t}{\lambda_t} \mathbf{G}_t^{\top}\mathbf{G}_t+\mathbf{I}_{D\times D}
\end{equation*}

\subsection{Learning and Inference}\label{sect:learning}

We learn the parameters in $\mathbf{\Theta}_t$ by marginal likelihood optimization. We switch between the primal and dual forms of the log marginal likelihood for computational efficiency and numerical stability. In its primal form, the log marginal likelihood is given by the following,
\begin{equation*}\label{equ:marginal_primal}
    \begin{split}
    \mathcal{L}( \mathbf{\Theta}_t )= & -\frac{N}{2}\log (2\pi\beta_t^{-1}) - \frac{\beta_t}{2}\|\mathbf{y}_t\|^{2} \\
    &+ \frac{\beta_t^2}{\lambda_t}\mathbf{y}_t^{\top}\mathbf{G}_t \mathbf{K}_{\mathbf{w}_t}^{-1} \mathbf{G}_t^{\top}\mathbf{y}_t - \frac{1}{2}\log|\mathbf{K}_{\mathbf{w}_t}|
    \end{split}
\end{equation*}
In its dual form, the log marginal likelihood is given as the logarithm of $\mathcal{N}(\mathbf{y}_t;\lambda_t^{-1}\mathbf{G}_t\mathbf{G}_t^{\top}+\beta_t^{-1}\mathbf{I}_{N\times N})$. When $N>D$, we optimize the log marginal likelihood in primal form, otherwise in dual form. Now, our operating assumption is that the black-box functions $f_1,...,f_T$ are \textit{correlated} in some way; accordingly, we let the neural network parameters be \textit{shared} over all the time steps $t$, giving the multi-task marginal likelihood \cite{perrone2018scalable},
\begin{equation}\label{equ:mtlobj}
\mathcal{L}(\mathbf{\Theta}) = \sum_{t=1}^{T} \mathcal{L}( \mathbf{\Theta}_t)
\end{equation}
where we have used $\mathbf{\Theta}$ to indicate $\cup_{t=1}^T\mathbf{\Theta}_t$. The overall computational complexity of optimizing $\mathcal{L}(\mathbf{\Theta}) $ is $O(T\max\{N,D\}(\min\{N,D\})^2)$. If $T$ is very large, then we can first randomly sample a subset of time steps $S\subset\{1,...,T\}$, and then maximize $\sum_{t\in S}\mathcal{L}( \mathbf{\Theta}_t)$ instead at each iteration of marginal likelihood optimization.

\textbf{Acquisition Function}. To construct the acquisition function $a_{f,t}(\mathbf{x}_t^{_\dagger}|\mathcal{A}_t)$ in BO (for test data point $\mathbf{x}_t^{_\dagger}$), we first obtain the feature map $\mathbf{g}_t$ by passing $\mathbf{x}_t^{_\dagger}$ through the neural network. Then the predictive distribution is obtained by integrating out $\mathbf{w}_t$ in the delta measure $\delta(f_t(\mathbf{x}_t^{_\dagger})  = \mathbf{w}_t^{\top}\mathbf{g}_t)$ with respect to its posterior in (\ref{eq:h_post}):
\begin{equation}\label{equ:posterior}
f_t(\mathbf{x}_t^{_\dagger}) \sim \mathcal{N}\big(\mu(\mathbf{x}_t^{_\dagger}|\mathcal{A}_{t},\mathbf{\Theta}_t),\sigma^2(\mathbf{x}_t^{_\dagger}|\mathcal{A}_{t},\mathbf{\Theta}_t)\big)
\end{equation}
where
\vspace{-1.0em}
\begin{equation}\label{equ:mean_var}
\begin{split}
\mu(\mathbf{x}_t^{_\dagger}|\mathcal{A}_{t},\mathbf{\Theta}_t) &= \mathbf{m}_{\mathbf{w}_t}^{\top}\mathbf{g}_t \\
\sigma^2(\mathbf{x}_t^{_\dagger}|\mathcal{A}_{t},\mathbf{\Theta}_t) &= \frac{1}{\lambda_t}\mathbf{g}_t^{\top}\mathbf{K}_{\mathbf{w}_t}^{-1}\mathbf{g}_t
\end{split}
\end{equation}

\vspace{-1em}
\begin{algorithm}[!b]
\caption{SMS-DKL}
\label{alg:main}
\begin{algorithmic}
\STATE {\bfseries Hyperparameters:} Max BO iterations $N$, max\\training iterations $M$, and acquisition function $a_{f}$
\STATE {\bfseries Input:} Sequence dataset $\mathcal{D}$
\STATE Initialize $\mathcal{A}_t$, $t\in\{1,...,T\}$ with random samples
\FOR{$n=1$ {\bfseries to} $N$}
\FOR{$m=1$ {\bfseries to} $M$}
\STATE Update $\mathbf{\Theta}_t$, $t\in\{1,...,T\}$ jointly\\~~~~by optimizing (\ref{equ:mtlobj})
\ENDFOR
\STATE Update $a_{f,t}(\mathbf{x}_t|\mathcal{A}_t)$, $t\in\{1,...,T\}$ using (\ref{equ:mean_var})
\STATE Solve $\mathbf{x}_{t}^{*}=\displaystyle\argmax_{\mathbf{x}_t\in \mathbb{X}}a_{f,t}(\mathbf{x}_t|\mathcal{A}_t)$, $t\in\{1,...,T\}$
\STATE Sample $\mathbf{x}^{*}$ from $\{\mathbf{x}_t^{*}: t\in\{1,...,T\}\}$\\~~~~via policy in (\ref{equ:policy})
\STATE $\mathcal{A}_{t}\leftarrow \mathcal{A}_t \cup (\mathbf{x}^{*},y_{t}^{*})$, $t\in\{1,...,T\}$
\ENDFOR
\STATE {\bfseries Output:} For each $t\in\{1,...,T\}$, the tuple $(\mathbf{x}_t,y_t)$\\with the best value of $y_t$ in $\mathcal{A}_{t}$
\end{algorithmic}
\end{algorithm}

\subsection{Acquisition Method}\label{sect:acquisition}

In standard BO, the next model to be acquired is chosen by maximizing the acquisition function $a_f(\mathbf{x}|\mathcal{A})$\textemdash e.g. the probability of improvement (PI) \cite{kushner1964new}, expected improvement \cite{jones1998efficient,mockus1978toward}, Gaussian process upper confidence bound (GP-UCB) \cite{srinivas2009gaussian}, and entropy search (ES) \cite{hennig2012entropy}:
\begin{equation}\label{equ:acq}
\mathbf{x}^{*} \in \argmax_{\mathbf{x}\in \mathbb{X}} a_f(\mathbf{x}|\mathcal{A})
\end{equation}
In DKL, the first two components produce the embedding vector $\mathbf{z}_t$. The third component is what takes $\mathbf{x}_t$ as input (along with $\mathbf{z}_t$), and its output is what the posterior in (\ref{equ:posterior}) and corresponding acquisition function are constructed with; we can optimize $a_{f,t}(\mathbf{x}_t,\mathcal{A}_t)$ by optimizing the input $\mathbf{x}_t$ only in the third network.

In our problem, we actually need to optimize $T$ black-box functions $f_1,...,f_T$ at the same time. In addition, recall that we can observe a given model's performance for \textit{all} time steps $t$ after every single model evaluation (i.e. we obtain $T$ data points $\{(\mathbf{x}_{n,t},y_{n,t})\}_{t=1}^{T}$ per acquisition). One straightforward solution is to simply define the sum $f_{\text{sum}}=\sum_{t=1}^T f_t$ and make acquisitions on the basis of this sum. However, this is not desirable since the optimizer of $f_{\text{sum}}$ is in general not identical to the optimizer of \textit{each} individual $f_t$. Optimizing $f_{\text{sum}}$ is only suitable if we were constrained to select a single model for all time steps $t$ (which we are not). In our case, at each BO iteration we first compute the \textit{individual} optimizers $\mathbf{x}_t^{*}$ for each acquisition function $a_{f,t}(\mathbf{x}_t|\mathcal{A}_t)$, $t\in\{1,...,T\}$. Then, our choice $c$ of which specific $\mathbf{x}_t^{*}$ to acquire is made via the following a stochastic policy, 
\begin{equation}\label{equ:policy}
p(c= \mathbf{x}_t^{*}|\mathbf{a})=\frac{a_{f,t}(\mathbf{x}_t^{*}|\mathcal{A}_t)}{\sum_{\tau=1}^{T}a_{f,\tau}(\mathbf{x}_\tau^{*}|\mathcal{A}_\tau)}
\end{equation}
where $\mathbf{a}=[a_{f,1}(\mathbf{x}_1^{*}|\mathcal{A}_1),...,a_{f,T}(\mathbf{x}_T^{*}|\mathcal{A}_T)]^{\top}$. This acquisition method is inspired by the Hedge algorithm  \cite{freund1999adaptive}. We treat each acquisition function as an expert giving advice as to which model to acquire. Assuming that we believe equally in all experts throughout the BO experiment, the probability of following the advice of the $t$-th expert is given by (\ref{equ:policy}).
Algorithm \ref{alg:main} provides pseudo- code summarizing our proposed method (SMS-DKL).

\section{Experiments and Discussion}\label{sect:experiments}

Sequence prediction admits a variety of models, among the most popular being RNNs in machine learning, although the difficulties of training them are widely recognized \cite{pascanu2013difficulty}. In this paper, we are motivated by the problem of temporal distribution shift within a dataset, a phenomenon especially relevant in the medical setting \cite{oh2019relaxed}, and in the presence of which the optimal model itself may vary over time. So far, we have \mbox{formalized} this challenge as one of stepwise model selection (SMS), and proposed a solution via deep kernel learning (DKL). Three questions remain, and our goal in this section is to answer them:
\vspace{-0.75em}
\begin{itemize}[leftmargin=*]
\itemsep0.12em
\item First, \textit{why} do we expect to benefit from stepwise selection at all? While the abstract notion of potential distribution shifts gives some intuition, here we empirically illustrate the validity of \mbox{SMS \textit{as the problem}}: We observe improvements simply by applying post-hoc stepwise selection over standard BO and MOBO.
\item  Second, \textit{what} is the practical benefit our technique for model selection? Here, we demonstrate the consistent, significant advantage of \mbox{DKL \textit{as the solution}}: We observe a clear improvement by addressing stepwise selection directly in the optimization procedure.
\item Third, \textit{how} do the correlations ultimately influence the optimal models selected? Here, we visualize the correlations in model performance over time, as well as the learned embeddings $\mathbf{z}_t$ and optimizers $\mathbf{x}_t$, shedding further light on the workings of SMS-DKL.
\end{itemize}

\vspace{-0.5em}
\textbf{Datasets}. We use three datasets in our experiments. The first consists of patients enrolled in the \href{https://www.cysticfibrosis.org.uk/the-work-we-do/uk-cf-registry}{UK Cystic Fibrosis registry} (\textbf{UKCF}), which records annual follow- up trajectories for over 10,000 patients in \mbox{2008--2015}.
At each time step, we issue predictions on the basis of 90 temporal variables (e.g. treatments, comorbidities, infections), focusing on
three important clinical outcomes (see e.g. \cite{alaa2019pass}): the 1-year mortality (1YM), allergic broncho-pulmonary aspergillosis (ABPA), and the lung infection E. coli. The second consists of patients in intensive care units from the \href{https://mimic.physionet.org}{MIMIC-III database} (\textbf{MIMIC}),
containing
physiological data streams for over 22,000 patients. During the first 48 hours of each episode, we issue predictions
using 40 temporal variables
(including the most frequently measured vital signs and lab tests)
focusing
on three important clinical outcomes (see e.g. \cite{oh2019relaxed}): acute respiratory failure (ARF), shock, and in-hospital mortality (IHM). The third (\textbf{WARDS}), assembled by \cite{alaa2017personalized}, consists of over 6,000 patients hospitalized in the general medicine floor of a major medical center in 2013--2015. On the basis of 21 physiological data streams (including vital signs and lab tests), we predict whether each patient will be admitted to critical care within 24 hours from the current time as a result of clinical deterioration (ICU).

\renewcommand{\arraystretch}{1.1}
\setlength\tabcolsep{6.65pt}
\begin{table*}[b]
\centering
\begin{small}
\begin{tabular}{l|ccc|ccc|c}
\toprule
\textit{Dataset} & \multicolumn{3}{c|}{UKCF}  & \multicolumn{3}{c|}{MIMIC} &  WARDS \\
\midrule
\textit{Target} & 1YM & ABPA & E. coli & IHM & Shock  & ARF & ICU \\
\bottomrule
\addlinespace[0.2em]\multicolumn{8}{c}{\textit{100 BO Iterations}}\\
\toprule
\textsc{GP}  & .586  $\pm$ .001 & .627  $\pm$ .001 & .872  $\pm$ .001 &.460  $\pm$  .003 &.107  $\pm$ .001 & .114   $\pm$ .001 &.161  $\pm$ .004   \\
GP$_{\textsc{WISE}}$ & .593  $\pm$ .002 & .636  $\pm$ .001 & .877  $\pm$ .001  & .463   $\pm$ .002 & .113  $\pm$  .001  &  .126   $\pm$  .001 & .168  $\pm$ .005  \\
{ParEGO}& .588   $\pm$ .002& .623   $\pm$ .002 & .872   $\pm$ .002 & .461 $\pm$ .001  & .107   $\pm$ .000  &  .118   $\pm$ .001 & .176  $\pm$ .003   \\
ParEGO$_{\textsc{WISE}}$& .594   $\pm$ .001& .633   $\pm$ .003 & .876   $\pm$ .001 & .464 $\pm$ .001  & .112  $\pm$ .001  &  .124   $\pm$ .001 & .184  $\pm$ .004   \\
{PESMO} & .592  $\pm$ .002   & .629 $\pm$ .001 & .874 $\pm$ .001  & .467  $\pm$ .001  & .111  $\pm$ .002 & .111  $\pm$ .002  & .176  $\pm$ .006    \\
PESMO$_{\textsc{WISE}}$ & .598  $\pm$ .002   & .639 $\pm$ .001 & .878 $\pm$ .001  & .469  $\pm$ .001  & .115  $\pm$ .001 & .123  $\pm$ .001  & .182  $\pm$ .005    \\
\textsc{SMS-DKL} & \bftab{.601}  $\pm$ \bftab{.001}& \bftab{.641}   $\pm$ \bftab{.001} & \bftab{.880}  $\pm$ \bftab{.001} & \bftab{.474} $\pm$ \bftab{.002}  & \bftab{.118}  $\pm$ \bftab{.001}  & \bftab{.128}  $\pm$ \bftab{.001} & \bftab{.197}  $\pm$ \bftab{.004}   \\
\bottomrule
\addlinespace[0.2em]\multicolumn{8}{c}{\textit{500 BO Iterations}}\\
\toprule
\textsc{GP}  & .592  $\pm$ .001 & .633  $\pm$  .002 &.876  $\pm$ .000 &.471  $\pm$  .001 &.113  $\pm$ .001 & .120   $\pm$ .001 &.176  $\pm$ .003   \\
GP$_{\textsc{WISE}}$ & .601  $\pm$ .001 & .644  $\pm$ .001 &  .882  $\pm$ .000  & .475  $\pm$ .001 & .122  $\pm$ .001  &  .135   $\pm$  .001 & .189  $\pm$ .002  \\
{ParEGO}& .593   $\pm$ .002& .632   $\pm$ .001 & .875   $\pm$ .001 & .469 $\pm$ .001  & .114   $\pm$ .001  &  .122   $\pm$ .002 & .187  $\pm$ .004   \\
ParEGO$_{\textsc{WISE}}$& .601  $\pm$ .001& .644  $\pm$ .001 & .881  $\pm$ .001 & .473  $\pm$ .001  & .121  $\pm$ .001 &  .136   $\pm$ .001 & .199   $\pm$ .002   \\
{PESMO} & .592  $\pm$ .001   & .632 $\pm$ .002   & .876 $\pm$ .000  & .469  $\pm$ .001  & .113  $\pm$ .002 & .117  $\pm$ .002  & .186  $\pm$ .004    \\
PESMO$_{\textsc{WISE}}$ & .601  $\pm$ .001   & .642  $\pm$ .001  & .881  $\pm$ .000  & .474 $\pm$ .001   &  .119  $\pm$ .001  &  .132 $\pm$ .001 &  .194  $\pm$ .002   \\
\textsc{SMS-DKL} & \bftab{.603}   $\pm$ \bftab{.001}& \bftab{.645}   $\pm$ \bftab{.001} & \bftab{.884}  $\pm$ \bftab{.000} & \bftab{.476}  $\pm$ \bftab{.001}  & \bftab{.123}   $\pm$ \bftab{.001}  &  \bftab{.138}  $\pm$ \bftab{.001} & \bftab{.207} $\pm$ \bftab{.002}   \\
\bottomrule
\end{tabular}
\vskip -0.05in
\caption{Performance of SMS-DKL and Benchmarks: AUPRC scores at the $100$-th and $500$-th BO iteration mark.}
\vskip -0.1in
\label{tab:realdata}
\end{small}
\end{table*}

\vspace{-0.1em}
\textbf{Experimental Setup.} We have a total of 7 sequence prediction tasks from the three datasets. Sequences are of length 6 (at 1-year resolution) for UKCF, length 24 (at 2-hour resolution) for MIMIC, and length 24 (at 1-hour resolution) for WARDS. Benchmarks are implemented using the \texttt{GPyOpt} library or original source code. In SMS-DKL, the RNN component is implemented using LSTMs, the DeepSets component as a ReLU network (with an output embedding $\mathbf{z}_t$ or $\mathbf{z}_{t,m}$ of size 1), and the MLP component as a feedforward tanh network. See Appendix A for additional details on implementation. The underlying search space is the space of RNN models for sequence prediction, and hyperparameters considered include the learning rate, batch size, training epochs, hidden state size, input dropout rate, recurrent dropout rate, and the $\ell_2$-regularization coefficient. See Appendix B for additional details on hyperparameter space. In the presence of label imbalance in the data (common to the medical setting; see Table \ref{tab:corr}), we focus on optimizing the area under the precision-recall curve (AUPRC) as performance metric. BO is carried out to a maximum of $500$ iterations. Each experiment is repeated for a total of 10 times, each with a different random training and validation split. Each run of the experiment uses a different random seed to initialize the BO algorithms, and the same random seed is used for all algorithms at each run.

\vspace{-0.1em}
\textbf{Benchmarks.} We compare the proposed \textsc{SMS-DKL} with a standard BO algorithm and two MOBO algorithms {ParEGO} \cite{knowles2006parego} and {PESMO} \cite{hernandez2016predictive} on the SMS problem. While other MOBO algorithms are available (such as \textsc{EHI} \cite{emmerich2008computation}, \textsc{SMSEGO} \cite{ponweiser2008multiobjective}, and \textsc{SUR} \cite{picheny2015multiobjective}), they are \textit{not} scalable to problems with a large number of objectives. In particular, \textsc{SMSEGO} and \textsc{EHI} make acquisitions by computing the hyper-volume gain,
which is expensive in high dimensions (i.e. large numbers of objectives). Similarly, \textsc{SUR} is an extremely expensive criterion only feasible for 2 or 3 objectives at most, because it computes the expected decrease in the area under the probability of improving the hyper-volume. 

\begin{figure*}[t]
\vskip -0.15in
\subfloat[Correlation in black-boxes $f_t$]{
\begin{minipage}{0.295\textwidth}
\vspace{0.1cm}
\includegraphics[height=0.750\linewidth, width=0.85\linewidth]{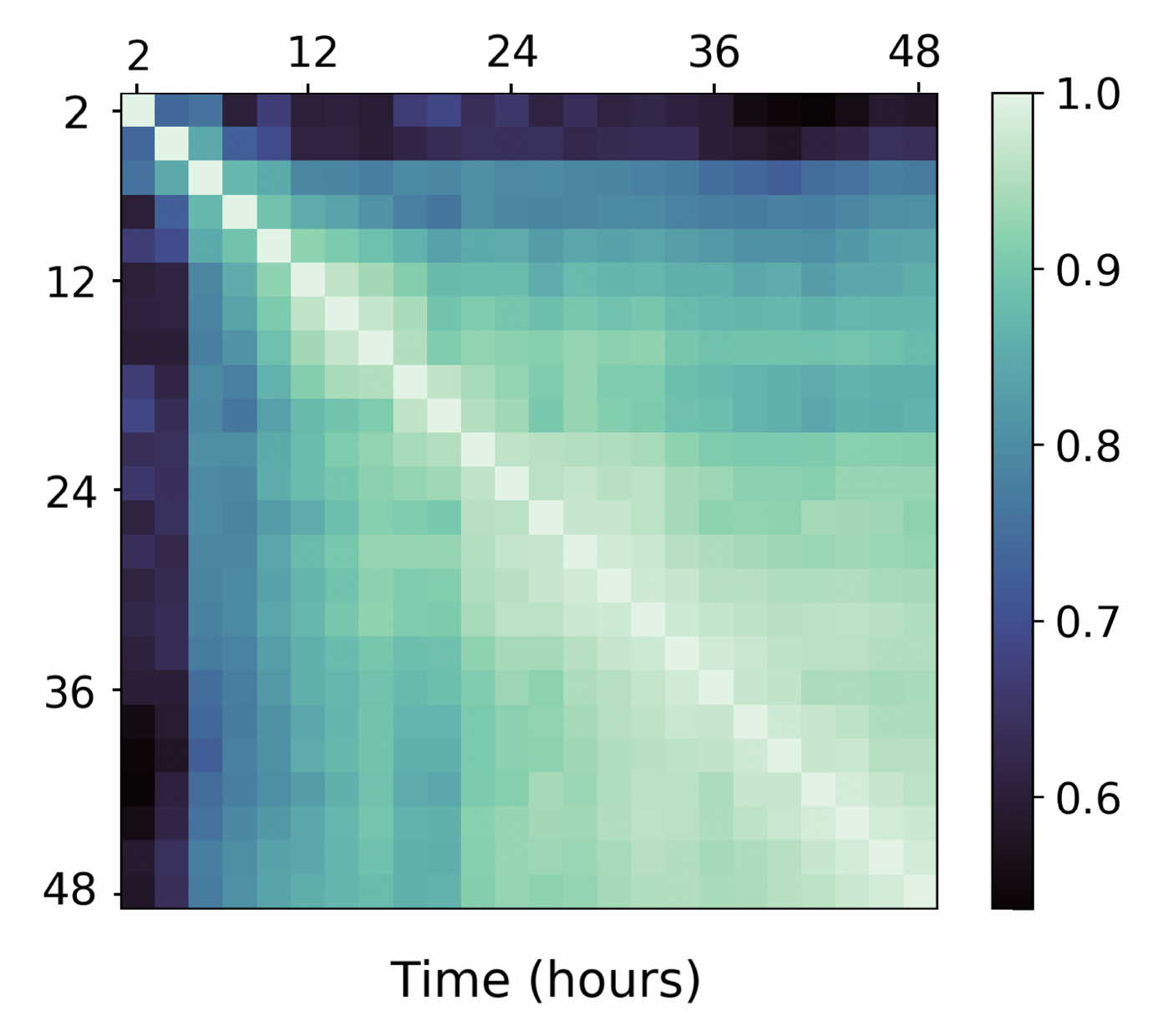}
\vspace{-0.18cm}
\label{fig:U1}
\end{minipage}}
\subfloat[Change in dataset embedding $\mathbf{z}_t$]{
\begin{minipage}{0.305\textwidth}
\includegraphics[height=0.7125\linewidth, width=0.92\linewidth]{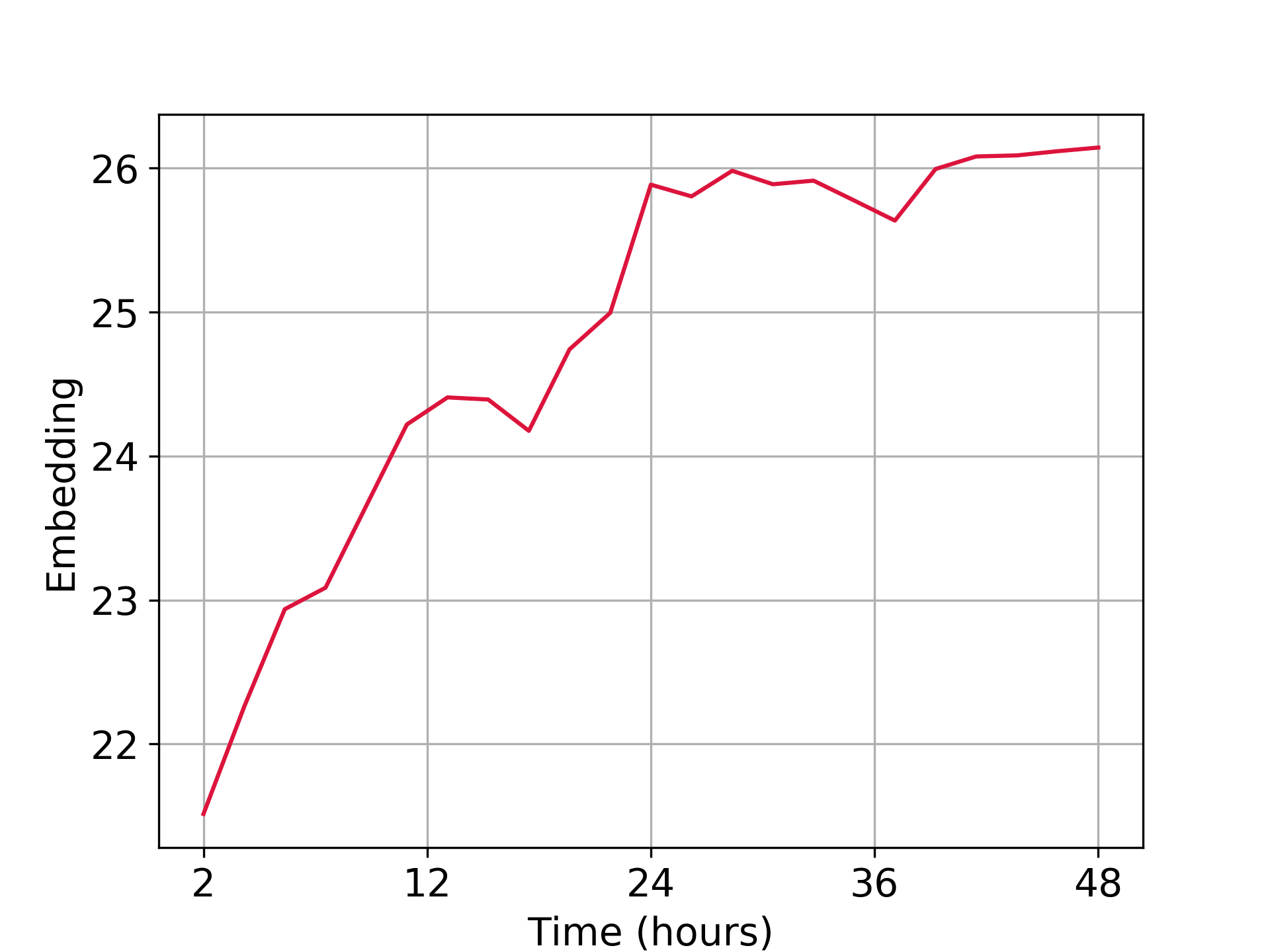}
\label{fig:U2}
\end{minipage}}
\subfloat[Change in optimal prediction model $\mathbf{x}_t$]{
\begin{minipage}{0.400\textwidth}
\includegraphics[height=0.5450\linewidth, width=0.9\linewidth]{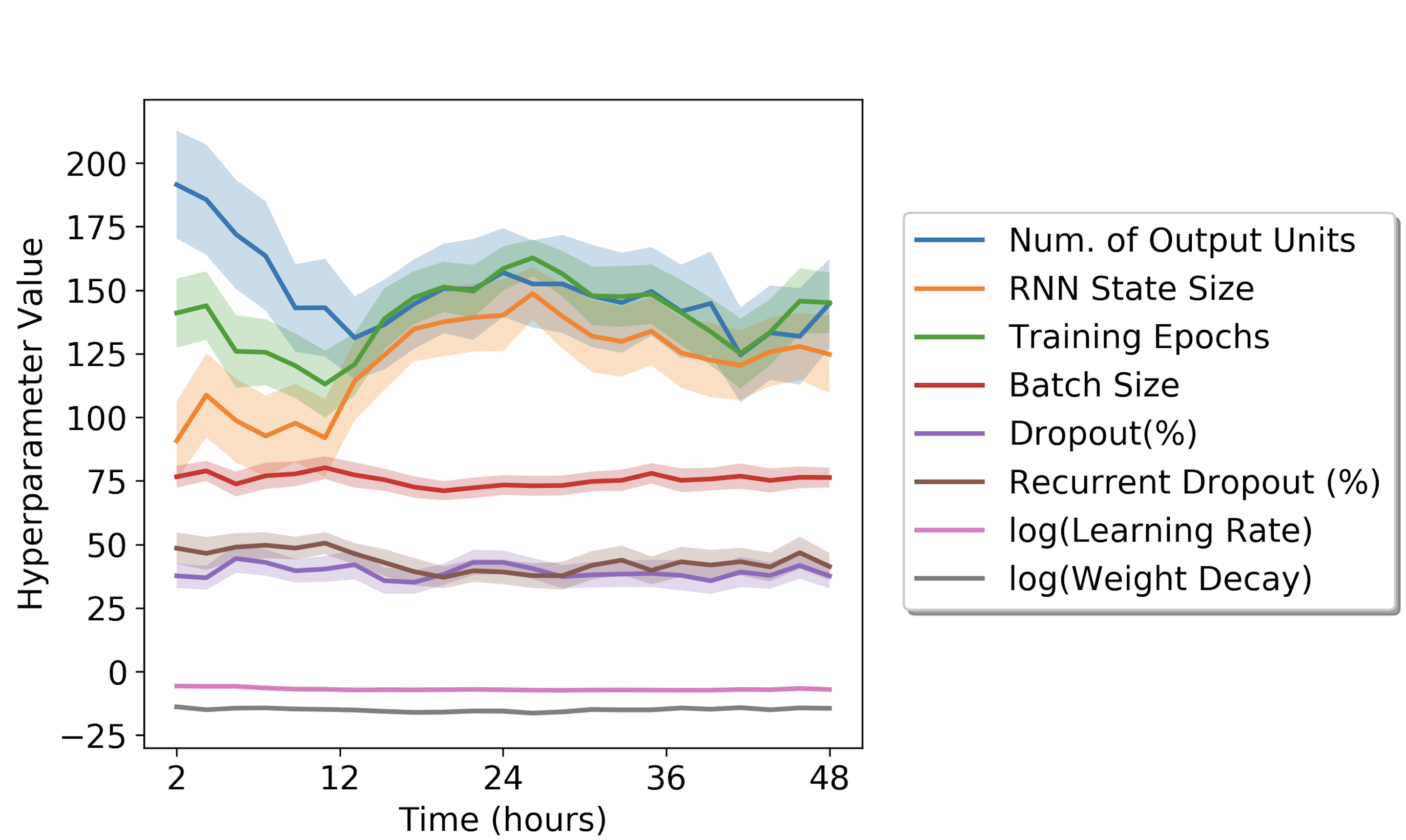}
\label{fig:U3}
\end{minipage}}
\vskip -0.05in
\caption{Learning correlations and changes over time. SMS-DKL leverages correlations to select stepwise-optimal models.}
\vskip -0.15in
\label{fig:model_performance}
\end{figure*}

\textbf{\textbullet} \textbf{GP}. We first consider a standard BO algorithm (GP) for comparison; this operates by simply optimizing the sum of the model performance metric over all steps $t$.

\textbf{\textbullet} \textbf{ParEGO} first transforms the multi-objective problem into a single-objective problem: At each BO iteration, the multiple objectives $f_t$ are scalarized into $f_{\bm{\theta}}$ using a randomly sampled weight vector $\bm{\theta} = (\theta_1,...,\theta_T)$,
\begin{equation}
f_{\bm{\theta}}(x) = \max_{t\in{T}}(\theta_t f_t(x)) +0.05\sum_{t=1}^{T}\theta_t f_t(x)
\end{equation}
Then at each BO iteration, a standard acquisition function can be used on $f_{\bm{\theta}}(x)$ to select the next point.

\setlength{\abovedisplayskip}{7.5pt}
\setlength{\belowdisplayskip}{6.0pt}

\textbf{\textbullet} \textbf{PESMO} is a recent, state-of-the-art MOBO algorithm based on predictive entropy search. The acquisition function in {PESMO} is expressed by the following,
\begin{equation}
a(x)= H(\mathcal{X}^{*}|\mathcal{D})-\mathbb{E}_y[H(\mathcal{X}^{*}| \mathcal{D}\cup \{(\mathbf{x},y)\})]
\end{equation}
where $H(\cdot)$ denotes the entropy and $\mathcal{X}^{*}$ is the Pareto set. {PESMO} operates by selecting the point that maximizes the information gain with respect to the Pareto set.

\textbf{\textbullet} \textbf{Post-hoc Stepwise Variants}. Since the goal in SMS is to attain good performance for \textit{each} individual time step, we additionally consider a straightforward modification to all aforementioned benchmarks that applies an extra post-hoc selection step, choosing (among all models) the best-performing model on a \textit{per-step} basis. For each of the algorithms considered (GP, ParEO, and PESMO), we denote by subscript ``\textsc{WISE}'' the results for this ``stepwise'' modification to the benchmark.

\subsection{Experiment Results}

\textbf{Overall Results}. For all BO algorithms considered, Table \ref{tab:realdata} reports the AUPRC score (averaged over all time steps $t$) at the $100$-th and $500$-th BO iteration. We now answer the first question posed at the beginning of this section: Is it reasonable to expect to benefit from SMS (at all)? Comparing each benchmark with its post-hoc stepwise modification, we answer in the affirmative: Across all benchmarks, observe that prediction performance is invariably improved simply by going back and selecting the best model on a per-step basis. The second\textemdash and perhaps more interesting\textemdash question is whether solving the SMS problem directly within the optimization procedure can offer additional gains. Comparing our proposed DKL solution with any comparator, the answer is also in the affirmative: Observe that SMS-DKL consistently and significantly outperforms both standard BO and MOBO algorithms across all datasets and prediction targets, at both the $100$-th and $500$-th BO iteration mark. This is true regardless of whether we allow comparators the additional freedom of choosing the best stepwise models post-hoc.

\textbf{Correlations and Changes}. Finally, how do changes and correlations over time ultimately play out with respect to optimal models? In SMS-DKL, the source of efficiency in optimization stems from its ability to learn the similarities and differences between black-box functions $f_t$. Using the prediction of shock in MIMIC as an example, Figure \ref{fig:U1} shows a correlation matrix of model performance across the 48-hour interval considered, at the $500$-th iteration mark. Here we see that black-box functions within first 12 hours are weakly correlated with subsequent times, and those within the latter half of the interval appear strongly correlated with each other\textemdash with values exceeding 0.95. Observe that this pattern is automatically picked up and reflected in the learned embeddings $\mathbf{z}_t$ of the datasets $\mathcal{D}_{\leq t}$ themselves: Figure \ref{fig:U2} shows these (one-dimensional) embeddings $\mathbf{z}_t$ over time. Notice a clear evolution consistent with the previous observation: the evolving datasets are autocorrelated, with particularly strong similarities among the latter 24-hour interval. Importantly, this plays out with respect to optimizers $\mathbf{x}_t$ in parallel: Figure \ref{fig:U3} shows the evolution of optimal models $\mathbf{x}_t$ over time; in order to highlight the trend across time steps, values are averaged over the 20 highest-performing models. (The conventional approach of selecting a single model would correspond to a series of flat lines). Consistent with our intuitions, notice\textemdash for instance\textemdash that models for earlier steps (which have access to less temporal information) appear to require less recurrent memory.

\begin{table}[b]
\vskip -0.15in
\begin{small}
\centering
\setlength\tabcolsep{4.2pt}
\begin{tabular}{c|c|c|cc}
\toprule
Dataset & Target & Autocorr. & \% Postive (start) & (end)  \\
\midrule
      & 1YM     & 0.592 & 12.2 & 23.4 \\
UKCF  & ABPA    & 0.564 & 27.6 & 12.7 \\
      & E. coli & 0.564 & 52.4 & 28.5 \\
\midrule
      & IHM     & 0.849 & 13.2 & 13.2 \\
MIMIC & Shock   & 0.867 &  8.9 &  8.9 \\
      & ARF     & 0.875 &  7.2 &  7.2 \\
\midrule
WARDS & ICU     & 0.976 &  1.7 &  3.2 \\
\bottomrule
\end{tabular}
 \vskip -0.05in
\caption{Feature autocorrelations and \% positive labels. The slight inter-task variation in feature autocorrelations within a dataset are due to label missingness and censoring.}
\label{tab:corr}
\vskip -0.025in
\end{small}
\end{table}

As an additional sanity check, Table \ref{tab:corr} also shows the first-order autocorrelations of input features in each dataset prediction task; these are first computed per feature, then averaged over all features. We see that the autocorrelations are stronger in MIMIC and WARDS than in UKCF. Although this is (at best) a rough proxy for the performance correlation between models across time steps, we observe an intuitive pattern: SMS-DKL shows more significant gains over datasets with higher autocorrelations. In particular, we outperform all of the benchmarks by the widest margin in WARDS.

\vspace{-0.25em}\subsection{Discussion}

The advantage of SMS-DKL for sequence prediction is predicated on the fact that the optimal model for predicting $e_t$ may change with $t$\textemdash within a given dataset. While there may be a variety of reasons for this phenomenon (encapsulated by the general notion of temporal distribution shift \cite{oh2019relaxed}), the benefit here is that optimal models are \textit{automatically} selected over time\textemdash agnostic as to the precise underlying mechanism of change, and \textit{without} requiring domain-specific engineering to explicitly model time-varying relationships.

\textbf{Generalization Performance}. Of course, a variety of sophisticated \textit{single-model} techniques can be\textemdash and have been\textemdash used to tackle temporal distribution shift; these include explicitly including temporal encodings, modeling abrupt transitions, mixing weights over time, as well as learning hypernetworks to modify the weights of the primary RNN model \cite{ha2016hypernetworks,oh2019relaxed}. On the one hand, such techniques have been shown to outperform the baseline RNN model on held-out test data; see \cite{oh2019relaxed} for a comprehensive analysis. On the other hand, extremely competitive (test set) performance can also be achieved via the simple application of SMS-DKL in optimizing the (unmodified) baseline RNN alone: In Appendix C, we show a head-to-head comparison of generalization performance for all such methods on MIMIC prediction tasks, and observe\textemdash interestingly\textemdash that SMS-DKL gives either the best or second-best test-set performance\textemdash purely by optimizing the baseline RNN model.


In this paper, we formalized the SMS problem in the context of sequence prediction, and developed the DKL algorithm as a solution. Using real-world datasets in healthcare, we illustrated the advantage of SMS-DKL over standard and multi-objective BO approaches for model selection. In contrast to alternative single-model techniques, we further verified the effectiveness of SMS-DKL with respect to generalization\textemdash with the added advantage that the method is simple and automatic.

\clearpage
\section*{Acknowledgements}
This work was supported by GlaxoSmithKline (GSK), Alzheimer's Research UK (ARUK), the US Office of Naval Research (ONR), and the National Science Foundation (NSF): grant numbers ECCS1462245, ECCS1533983, and ECCS1407712. We thank the reviewers for their helpful comments. We thank the UK Cystic Fibrosis Trust, the MIT Lab for Computational Physiology, and Ahmed M. Alaa respectively for making and/or preparing the UKCF, MIMIC, and WARDS datasets available for research.

\balance
\bibliography{sample_paper}
\bibliographystyle{IEEE}

\end{document}